\documentclass[10pt,twocolumn,letterpaper]{article}

\usepackage{iccv}
\usepackage{times}
\usepackage{epsfig}
\usepackage{graphicx}
\usepackage{amsmath}
\usepackage{amssymb}
\usepackage{booktabs}

\usepackage{multirow}
\usepackage{makecell}
\usepackage{subcaption}
\usepackage{url}

\makeatletter
\newcommand{\printfnsymbol}[1]{%
  \textsuperscript{\@fnsymbol{#1}}%
}
\makeatother

\usepackage[breaklinks=true,bookmarks=false]{hyperref}

\iccvfinalcopy 

\def\iccvPaperID{****} 

\ificcvfinal\fi

\begin{document}

\title{1st Place Solution for ICCV 2023 OmniObject3D Challenge: Sparse-View Reconstruction}

\author{Hang Du\thanks{These authors contributed equally to this project.}, Yaping Xue\printfnsymbol{1}, Weidong Dai, Xuejun Yan, Jingjing Wang \\
Key Laboratory of Peace-building Big Data of Zhejiang Province \\
\tt\small {duhang.shu@foxmail.com}
}

\maketitle
\ificcvfinal\thispagestyle{empty}\fi

\section{Introduction}
In this report, we present the 1st place solution of team ``\textbf{hri test}" to ICCV 2023 OmniObject3D Challenge Track-1 Sparse-View Reconstruction~\cite{challenge}. 
We ranked first in the final challenge test with a PSNR of 25.44614.  
The challenge aims to evaluate approaches for novel view synthesis and surface reconstruction using only a few posed images of each object. 
The number of input images varies from 1, 2 to 3, respectively.
Such scenario is a very challenging setting for novel-view synthesis and surface reconstruction. 
In the following, we will provide our solution and findings in the experiment. 

\section{Method}
\label{method}
\begin{figure}[t]
    \centering
    \includegraphics[scale=0.3]{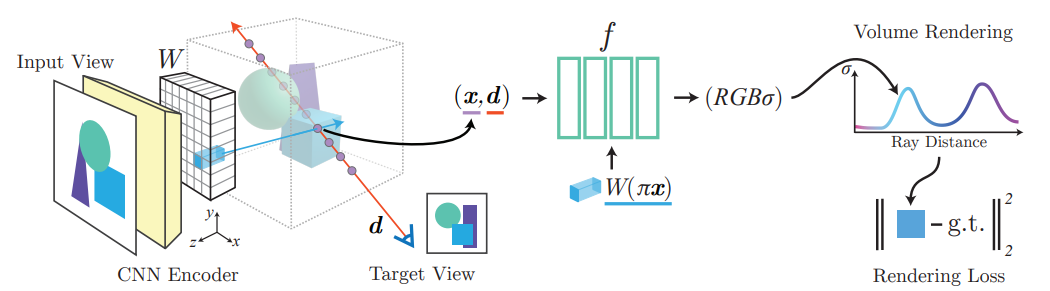}
    \caption{Overview architecture of  Pixel-NeRF~\cite{yu2021pixelnerf}. For a query point $x$ along a target ray with view direction $d$, they extract a corresponding feature from the image feature volume $W$ via projection and interpolation. Then, they feed the feature with spatial coordinates into the NeRF model, and predict RGB color values for neural rendering. }
    \label{pixel}
\end{figure}

In our solution, we choose NeRF-based methods as the fundamental approach for sparse-view reconstruction. 
Specifically, we pre-train Pixel-NeRF~\cite{yu2021pixelnerf} model to learn prior knowledge across different object categories from OmniObject3D training dataset~\cite{wu2023omniobject3d}, and then fine-tune the model on each test sense in a short period of time. 
Original NeRF~\cite{mildenhall2020nerf} model represents each scene as a function of colors and volume densities, optimizing the network on the dense views of every scene independently. 
To improve the generalization ability of NeRF models on few input images, Pixel-NeRF uses a local CNN to extract features from the input images. 
The overview architecture of Pixel-NeRF is shown in Figure~\ref{pixel}. One can refer to their original paper for more details.
Furthermore, we explore two additional strategies to  improve the performance of the Pixel-NeRF model on sparse-view reconstruction, including depth supervision, and coarse-to-fine positional encoding.

\subsection{Depth Supervision}
Recently, several methods~\cite{deng2022depth,yu2022monosdf,wang2023sparsenerf} develop depth constraints to improve the quality of surface reconstruction.   
Here, we use ground-truth depth maps to optimize the NeRF density function. 
To this end, we obtain the depth of each ray by 
\begin{equation}
d_{\mathbf{r}}=\sum_{i=1}^N w_i t_i, 
\end{equation}
where $w_i=T_i(1-\exp \left(-\sigma_i \delta_i\right))$, $T_i=\exp(-\sum_{j=1}^{i-1} \sigma_j \delta_j)$,  $\delta_i=t_i-t_{i-1}$, $\sigma_i$ is the predicted volume density from the network, and $t_i$ is the sampled point along the ray. Then, we directly compute MSE loss between the predicted depth $d_{\mathbf{r}}$ and ground-truth depth $\hat{d}_{\mathbf{r}}$ for all camera rays $\mathbf{R}$ of target pose $\mathbf{P}$ as 
\begin{equation}
\mathcal{L}_{\mathbf{depth}}=
\sum_{\mathbf{r} \in \mathbf{R}(\mathbf{P}) }\|d_{\mathbf{r}}-\hat{d}_{\mathbf{r}}\|^2. 
\end{equation}

Thus, the total training supervision consists of the rendered RGB pixel value and depth for each camera ray $\mathbf{r}$.

\subsection{Coarse-to-fine Positional Encoding}
In addition, certain studies~\cite{yang2023freenerf,lin2021barf,Truong2022SPARFNR} find that high-frequency inputs are prone to the over-fitting issue in sparse-view reconstruction. 
To tackle this problem, they propose a similar idea of gradually activating the high-frequency components of the positional encodings during the optimization process. By doing so, they aim to prevent NeRF models from immediately overfitting to the sparse-view training images. 
Thus, we follow the manner of BARF~\cite{lin2021barf} to assign weights to the $k$-th frequency component of positional encoding,
\begin{equation}
\gamma_k(\mathbf{x} ; \alpha)=w_k(\alpha) \cdot\left[\cos \left(2^k \pi \mathbf{x}\right), \sin \left(2^k \pi \mathbf{x}\right)\right], 
\end{equation}
where $w_k(\alpha)=$ is the weight function as 
\begin{equation}
w_k(\alpha)= \begin{cases}0 & \text { if } \alpha<k \\ \frac{1-\cos ((\alpha-k) \pi)}{2} & \text { if } 0 \leq \alpha-k<1 \\ 1 & \text { if } \alpha-k \geq 1\end{cases}
\end{equation}
where $\alpha \in[0, L]$ is a controllable parameter proportional to the iteration of training process, and $L$ is the number of frequency bands.
As the training process, such manner will activate more higher frequency bands of the encodings.

\section{Experiments}
In this section, we provide more implementation details and experimental results. 

\subsection{Data Preparation}
Following the guidelines of Track-1, we utilize OmniObject3D~\cite{wu2023omniobject3d} as the initial training dataset. 
OmniObject3D is a large-scale 3D object dataset, and thus requires an expensive cost when trained on the whole dataset. 
To reduce the training consumption, we select representative categories which share the same semantics or similar shape with the 80 test scenes. 
By doing so, we obtain about 48 training categories from the original full dataset. The details of 48 training categories can be found in the source code. 
Moreover, we also construct a validation set which consists of 80 test scenes from the above 48 training categories. This validation set allows us to evaluate the performance of our model locally.

\subsection{Implementation Details}
We implement Pixel-NeRF from the officially code~\cite{pixelnerf} in the experiment. 
For the image encoder, we use a ResNet34 backbone pre-trained on ImageNet and all categories in OmniObject3D training dataset. 
Then, we optimize the model for about 1.3 million  iterations on four P40 GPUs.
Finally, we fine-tune the pertained model on each test scene in 5K iteration. 
All training hyper-parameters are given in the source code.

\subsection{Results on Validation Set}

\begin{table}[t]
\begin{center}
\centering
\caption{Performance comparisons of two training datasets. }
\label{validation}
\resizebox{1\linewidth}{!}{
\begin{tabular}{c|ccc|ccc}
\toprule[1pt]

  \multirow{2}{*}{Method}
  &\multicolumn{3}{c|}{PSNR} 
  &\multicolumn{3}{c}{Chamfer Distance (CD)}  \\ 
 &{1-view }&{2-view }&{3-view }& {1-view  }&{2-view }&{3-view }\\
\midrule[0.5pt]
10 Categories &19.436&24.409&24.471&0.0393 &0.0349&0.0303\\ 
48 Categories  &\textbf{20.885}&\textbf{25.133}&\textbf{25.327}&\textbf{0.0388}
&\textbf{0.0321}
&\textbf{0.0282}
\\
\bottomrule[1pt]
\end{tabular}}
\end{center}
\vspace{-1em}
\end{table}

\textbf{Ablation of training data.}
We first provide the performance comparison under two training subsets in Table~\ref{validation}. 
The first line of the table corresponds to the model trained on the 10 categories with the most various scenes selected by OmniObject3D~\cite{wu2023omniobject3d}, while the second line represents training on the 48 representative categories mentioned earlier. From the results, we can observe that our representative categories provide abundant information for learning the object priors, and thus obtain better novel-view results.

\begin{table}[t]
\begin{center}
\centering
\caption{Ablation of additional model components.}
\label{Ablation}
\resizebox{1\linewidth}{!}{
\begin{tabular}{c|ccc|ccc}
\toprule[1pt]

 \multirow{2}{*}{Method}
 &\multicolumn{3}{c|}{PSNR} 
 &\multicolumn{3}{c}{Chamfer Distance (CD)}  \\ 
 &{1-view }&{2-view }&{3-view }& {1-view  }&{2-view }&{3-view }\\
\midrule[0.5pt]
baseline &19.436&24.409&24.471&0.0393 &0.0349&0.0303\\ 
\midrule[0.5pt]
+ DS &19.188&24.459&24.958&\textbf{0.0368} &0.0318&0.0280\\
+ C2FPE &18.610 &24.773
 &24.692&0.0398&0.0349&0.0304 \\
+ FT encoder &\textbf{19.757} &24.490
 &24.687&0.0390 &0.0360 &0.0307\\
\midrule[0.5pt]
Full model &19.304&\textbf{24.877}
 &\textbf{25.119}&0.0384 &\textbf{0.0312} &\textbf{0.0275}
 \\

\bottomrule[1pt]
\end{tabular}}
\end{center}
\vspace{-1em}
\end{table}

\textbf{Ablation of model components.}
We also conduct ablation study on each model component. As shown in Table~\ref{Ablation}, the ``DS'' denotes the extra depth supervision, the ``C2FPE'' represents the coarse-to-fine positional encoding method, and ``FT encoder'' is to fine-tune the image encoder on all categories in OmniObject3D training dataset. 
The ``Full model'' refers to the model that incorporates all the model components.
Moreover, the depth supervision is effective to predict more accurate depth and improve the density field modeling, resulting in a better fidelity of surface reconstruction.

\begin{table}[t]
\begin{center}
\centering
\caption{Study of different test-time optimization strategies. }
\label{test_time}
\resizebox{1\linewidth}{!}{
\begin{tabular}{c|ccc|ccc}
\toprule[1pt]

  \multirow{2}{*}{Method}
  &\multicolumn{3}{c|}{PSNR} 
  &\multicolumn{3}{c}{Chamfer Distance (CD)}  \\ 
 &{1-view }&{2-view }&{3-view }& {1-view  }&{2-view }&{3-view }\\
\midrule[0.5pt]
w/o fine-tuning &21.645&23.362&24.757&0.0394 &0.0339&0.0293\\ 
\midrule[0.5pt]
freeze encoder
 &25.081&\textbf{26.249}&28.603&\textbf{0.0344}
&\textbf{0.0264}
&0.0210
\\
freeze rendering &23.279 &25.829
 &\textbf{28.937}&0.0387 &0.0277 &\textbf{0.0202}\\ 
Full fine-tuning &\textbf{25.116} &25.881
 &28.679&0.0353 &0.0279&0.0207\\

\bottomrule[1pt]
\end{tabular}}
\end{center}
\vspace{-1em}
\end{table}

\textbf{Study of test-time optimization.}
Furthermore, we try different object-level fine-tuning strategies. Specifically, we freeze the encoder or rendering module during the test-time optimization. 
Note that the model is only evaluated on the half  of validation set, due to the limited computation resources.
From the results in Table~\ref{test_time}, we can conclude that object-level fine-tuning is crucial in improving the quality of sparse-view reconstruction, particularly for unseen scenes.
This optimization step helps the model adapt to the specific characteristics of each test scene, resulting in better reconstruction results.

\subsection{Results on Challenge Test}
\begin{figure}[t]
    \centering
    \includegraphics[scale=0.425]{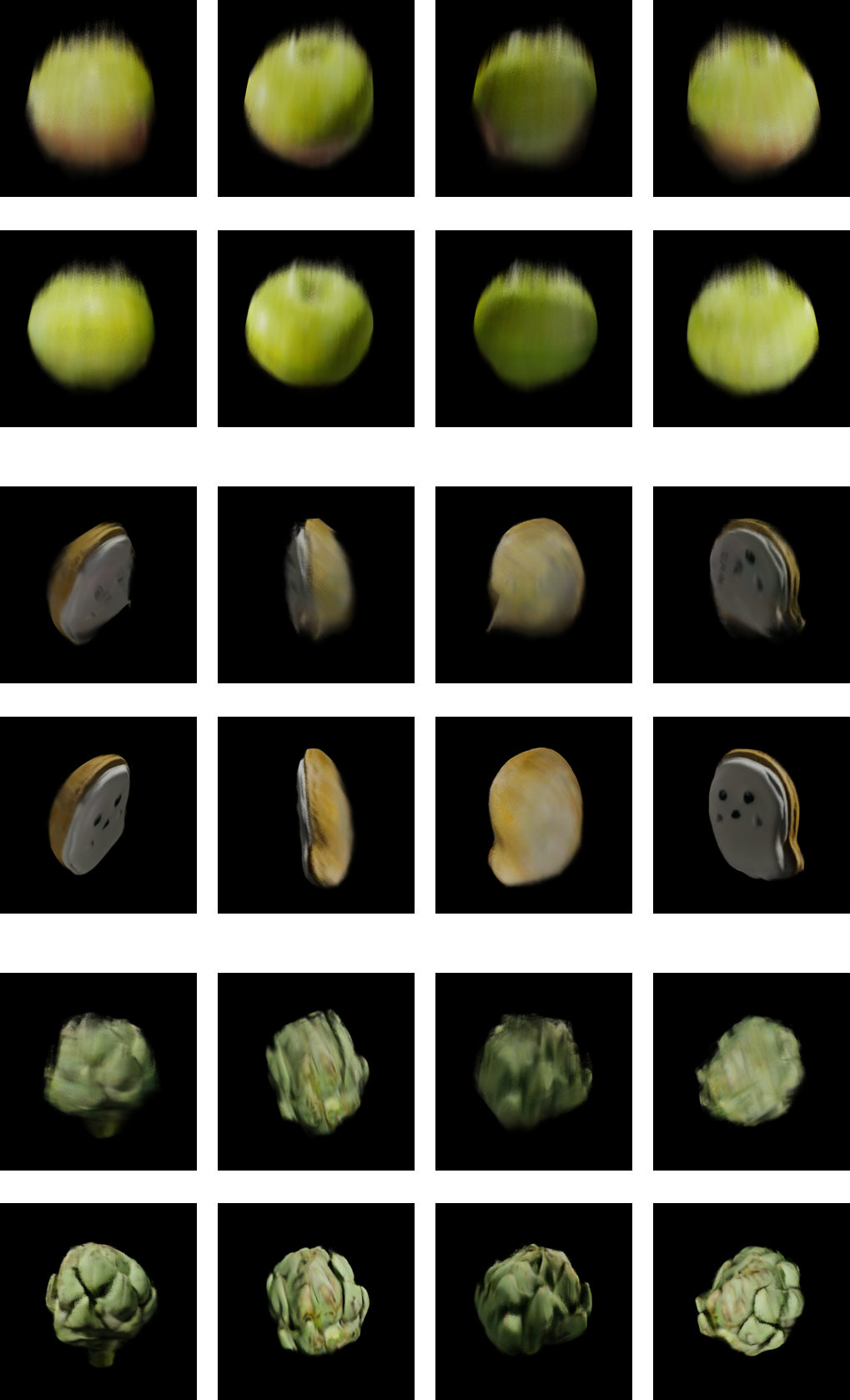}
    \caption{Qualitative comparisons on challenge test. The first line of each object is generated by the pre-trained model, and the second line is generated by each object-level fine-tuned model. }
    \label{challenge_vis}
\end{figure}

Here, we report more results on the challenge test. 
As mentioned before, we pre-train the Pixel-NeRF model on the representative 48 categories, and then fine-tune the model on each test scene.
The results are presented in Table~\ref{challenge}.
Moreover, as shown in Figure~\ref{challenge_vis}, we can clearly observe improved results achieved by the fine-tuning process. 

\begin{table}[t]
\begin{center}
\centering
\caption{Performance comparisons on challenge test. }
\label{challenge}
\resizebox{0.7\linewidth}{!}{
\begin{tabular}{c|c|c}
\toprule[1pt]
  \multirow{1}{*}{Method}
  &\multicolumn{1}{c|}{PSNR} 
  &\multicolumn{1}{c}{CD}  \\ 
\midrule[0.5pt]
w/o fine-tuning &23.32744&0.03268
\\
w fine-tuning  &\textbf{25.44614}&\textbf{0.02794}\\
\bottomrule[1pt]
\end{tabular}}
\end{center}
\vspace{-1em}
\end{table}

\section{Conclusion}
In this report, we introduce the details of our solution to ICCV 2023 OmniObject3D Challenge Track-1. We utilize Pixel-NeRF as the basic model, and apply depth supervision as well as coarse-to-fine positional encoding. 
The experiments demonstrate the effectiveness of our approach in improving sparse-view reconstruction quality. We ranked first in the final test with a PSNR of 25.44614.

{
\bibliographystyle{ieee_fullname}

}

\end{document}